\documentclass{article}
\usepackage[utf8]{inputenc}
\usepackage{listings}
\usepackage{url}
\usepackage{geometry}
\usepackage{color}
\usepackage{hyperref}
\usepackage{easy-todo}
\usepackage{graphicx}
\usepackage[dvipsnames]{xcolor}
\usepackage{mathtools}

\begin{document}
\vspace*{0.2in}

\begin{flushleft}
{\Large
\textbf\newline{CTNN: Corticothalamic-inspired neural network}
}
\newline

Leendert A Remmelzwaal \textsuperscript{1*},
Amit K Mishra \textsuperscript{1},
George F R Ellis \textsuperscript{2}
\\
\bigskip
\textbf{1} Department of Electrical Engineering, University of Cape Town, Rondebosch, Cape Town,
South Africa 7700.
\\
\textbf{2} Department of Mathematics and Applied Mathematics, University of Cape Town, Rondebosch, Cape Town, South Africa 7700.
\\
\bigskip

* Corresponding author: leenremm@gmail.com

\end{flushleft}

% =============================================================== %

\section*{Abstract}

Sensory predictions by the brain in all modalities take place as a result of bottom-up and top-down connections both in the neocortex and between the neocortex and the thalamus. The bottom-up connections in the cortex are responsible for learning, pattern recognition, and object classification, and have been widely modelled using artificial neural networks (ANNs). Here, we present a neural network architecture modelled on the top-down corticothalamic connections and the behaviour of the thalamus: a corticothalamic neural network (CTNN), consisting of an auto-encoder connected to a difference engine with a threshold. We demonstrate that the CTNN is input agnostic, multi-modal, robust during partial occlusion of one or more sensory inputs, and has significantly higher processing efficiency than other predictive coding models, proportional to the number of sequentially similar inputs in a sequence. This increased efficiency could be highly significant in more complex implementations of this architecture, where the predictive nature of the cortex will allow most of the incoming data to be discarded.

% =============================================================== %

\section*{Glossary}

\noindent{ANN = artificial neural network}\\
\noindent{AR = augmented reality }\\
\noindent{CTNN = corticothalamic neural network}\\
\noindent{LGN = lateral geniculate nucleus}\\
\noindent{MGB = medial geniculate body}\\
\noindent{MSE = mean square error}\\
\noindent{RTN = reticular nucleus}\\
\noindent{VB = ventrobasal complex}

% =============================================================== %

\section{Introduction}

In this paper we model the way sensory data is pre-processed by the thalamus in the human brain before being sent to the neocortex, thereby enabling major savings in processing power in the cortex, and also enabling filling in of occluded data. The key point which our model focuses on is stated by Andy Clark \cite{clark2015embodied} in discussing Radical Predictive Processing: ``\textit{The core flow of information is top-down, not bottom-up, and the forward flow of sensory information is replaced by the forward flow of prediction error}." Apart from the gustatory system, all sensory inputs reach the cortex only via the thalamus \cite{alitto2003corticothalamic}; that is where the prediction error signal is generated. The Corticothalamic Neural Network (CTNN) architecture presented here is part of the structure by which the predictive brain handles sensory processing in ways that significantly enhance survival prospects - after all, these structures have been hardwired into the nervous system by evolutionary processes precisely because they provide such an advantage. This suggests that the principles demonstrated here, which apply equally to visual, audio, and somatosensory data, may well play a useful role in both AI and robotics applications. 

% =============================================================== %

\subsection{The neocortex and corticothalamic connections}

\begin{figure}[ht!]
\centering
\includegraphics[scale=0.55]{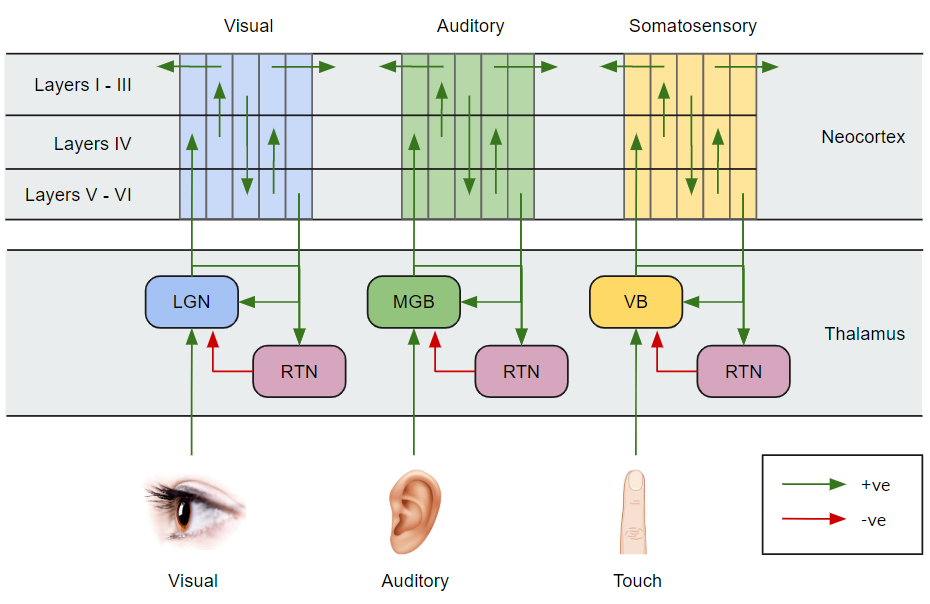}
\caption{\textit{Corticothalamic circuitry for the (a) visual, (b) auditory, and (c) somatosensory systems, adapted from Alitto and Usrey \cite{alitto2003corticothalamic}. Sensory information first enters the thalamus (the LGN in the case of vision) before proceeding to the functional hierarchy of cortical regions responsible for processing that specific sensory signal (e.g. V1 and V2 in the case of vision) \cite{ungerleider1982two} \cite{orban2004comparative} \cite{goodale1992separate} \cite{hegde2007reappraising} \cite{perry2014feature}. From the thalamus, information is sent to the cortex \cite{alitto2003corticothalamic} \cite{sillito2006always} \cite{hegde2007reappraising} \cite{hawkins2007intelligence}. Feedback also occurs between the lowest regions of the functional hierarchy back to the thalamus, specifically the reticular nucleus (RTN) \cite{alitto2003corticothalamic}. \vspace{5mm}}}
\label{fig:Fig13}
\end{figure}

The neocortex is a thin sheet of neural tissue enveloping most of the older parts of the mammalian brain, measuring only 2.5mm thick and functionally connected as an extensive hierarchy of cortical columns \cite{mountcastle1997columnar} \cite{clark2013whatever} \cite{burnod1990adaptive}. The thalamus acts as a relay station between visual, auditory and somatosensory sensory inputs and the neocortex, modulating the input based on previously processed information \cite{alitto2003corticothalamic} \cite{miller2002spectrotemporal} \cite{sillito2002corticothalamic} \cite{briggs2008emerging}. The connections between the thalamus and the neocortex are referred to as corticothalamic connections \cite{sherman2009exploring} and consist of both bottom-up (feed forward) and top-down (feedback) connections \cite{ellis2018top}. These corticothalamic connections play an important role in cognition \cite{jones2002thalamic}, attention direction \cite{shipp2004brain} \cite{de2014thalamic} \cite{wimmer2015thalamic} \cite{friston2018does}, sensory selection \cite{thalamic_saalmann2014neura_attentionl} \cite{thalamic_control_attention} \cite{thalamic_zikopoulos2007circuits}, awareness \cite{rees2009visual}, emotional control \cite{sun2015human}, prediction \cite{ellis2017beyond}, and salience detection \cite{bowman2013attention}. In 2003 Alitto and Usrey \cite{alitto2003corticothalamic} published a seminal paper describing the corticothalamic circuitry for the visual, auditory, and somatosensory systems (see Figure \ref{fig:Fig13}). In this paper there are two key observations we reference: (1) all three sensory systems (visual, auditory, and somatosensory) share a similar basic modular structure: a generalized processing unit which we call a ``generalized computational unit" in this paper, and (2) while the thalamus receives input sensory systems, it also receives top-down (feedback) connections from the cortex by means of the reticular nucleus (RTN). The thalamus is then responsible for comparing the incoming signal to the top-down prediction signal, and sending a moderated signal back to the cortex if the difference between the predicted signal and incoming data exceeds a threshold (see Figure \ref{fig:Fig13}). 

\subsection{Key points}

The central point we focus on is that there is no direct route from the eye or ear or somatory senses to the cortex: they all go through the thalamus, with essentially the same architecture in all three cases \cite{alitto2003corticothalamic} (see Figure \ref{fig:Fig13}). That is where expectations of what should be observed are compared with incoming data - what is actually observed - and a moderated signal is sent up to the cortex as the only input from the actual data to the cortex for higher level processing. Thus the thalamus acts on the incoming data as a difference engine between expectations and what is detected. Another key point is that adding the thalamus as a lower level is not equivalent to just adding another layer to the cortex, as its logical structure is quite different than that of adjacent cortical layers \cite{alitto2003corticothalamic}. In particular, it does not store data for future use by altering connection weights, as the cortex does. Instead, the thalamus contains relay cells influenced by feedback connections from the cortex \cite{sillito2002corticothalamic} \cite{alitto2003corticothalamic}. This architecture allows a very effective and computationally efficient form of predictive coding - by reducing processing at higher levels, the human brain is allowed to process other tasks (e.g. allowing the driver of the car to have a conversation while simultaneously driving). In addition, this architecture also demonstrates robustness when affected by occlusions of one of more of the senses.

% =============================================================== %

\subsection{Scope of this paper}

In this paper, using a cognitive architecture approach \cite{Cog-Architectures_lieto2018role}, we propose a generalized computational model of bottom-up and top-down connections in the neocortex, and bottom-up and top-down connections between the thalamus and the cortex, capable of simultaneously processing different inputs types.

\vspace{5mm}

\noindent{}The proposed generalized computational model should demonstrate the following:

\begin{enumerate}
\item High processing efficiency, specifically when processing sequences of sequentially similar inputs (e.g. images).
\item Input type agnostic - able to accept various input types including auditory, visual and somatosensory.
\item Multi-modal - able to process multiple input types simultaneously, combining them to an emergent realization. 
\item Robustness when there is partial occlusion of one or more sensory inputs \cite{purves2010brains}.
\end{enumerate}

% =============================================================== %

\subsection{The basic underlying assumptions}

There are two basic assumptions underlying this paper, as in the companion SANN paper \cite{remmelzwaal2020salience}. Firstly, evolution has fine-tuned human brain structure over millions of years to give astonishing intellectual capacity. It must be possible for designers of ANNs to learn possible highly effective neural architectures from studying brain structure \cite{edelman2007learning}. This has of course happened in terms of the very existence of ANNs, based in modelling the structure of cortical columns \cite{mountcastle1997columnar} \cite{clark2013whatever}. It has not happened as regards the structure and function of the corticothalamic systems in the way presented here, based in the work of Alitto and Usrey (below we review models that are broadly similar but in fact different) \cite{alitto2003corticothalamic}. However they have been hardwired into the human brain precisely because they perform key functions that have greatly enhanced survival prospects. The way this architecture works should therefore have the capacity to increase performance of any kind of ANN, and so has the potential to play a significant role in robotics or AI. 

Secondly, while the brain is immensely complex and therefore requires study at all scales of detail in order that we fully understand it, nevertheless there are major basic principles that characterise its structure and function, that can be very usefully developed in simple models such as presented here. These can provide an in-principle proof that the concept works, and so may be worth incorporating in much more complex models such as massive deep learning or reinforcement learning networks. The testing we do on the simple models described here suggests that may indeed be the case.

% =============================================================== %

\subsection{Limitations}

To the authors' knowledge, this is the first work in the open literature where an architecture has been proposed inspired by the corticothalamic interaction. We have only shown some results from limited experiments. Many more experiments will be needed to show the strength and weaknesses in a variety of situations, for example involving face and voice recognition, and situations such as driving down a road.

The CTNN model we present does not make use of recurrent connections, and therefore we do not take into account any temporal components to a series of images during training and testing. This may have applications in correctly reading text that has been garbled. This model could be further extended by addition of many more layers \cite{nielsen2015neural} \cite{lin2017does}. It is also worth noting that CTNNs do not currently handle the effects of salience signals, which may help us understand how we are able to directing attention when driving down a road. This will be the subject of future work, which will relate these attentional effects to salience signals which set the triggering threshold. Also, we do not attempt to model spiking neurons in the CTNN model; that again can be the subject of future investigation.

% =============================================================== %

\subsection{Related work}

In this section, we consider the relation of our model to other models.

\subsubsection*{Predictive coding models}

The current computational models that are similar to what we present here in some ways are predictive coding models. Predictive coding models are computational architectures which include both bottom-up and top-down connections, allowing for bio-realistic predictions of the current context to be fed back down to the sensory inputs by higher cognitive regions. There have been many distinctive predictive coding architectures developed over the last 20 years, including:

\vspace{5mm}

    \begin{center}
    \begin{tabular}{ |p{8cm}|p{3cm}|p{1cm}|  }
        \hline
        \multicolumn{3}{|c|}{Predictive coding models} \\
        \hline
        Model Name&Primary Author&Date\\
        \hline
        Retinal Predictive Coding \cite{srinivasan1982predictive}&Srinivasan&1982 \\
        \hline
        Linear Predictive Coding \cite{o1988linear}&O’Shaughnessy&1988 \\
        \hline
        Cortical Predictive Coding \cite{rao1999predictive}&Rao and Ballard&1999 \\
        \hline
        Restricted Boltzmann machine (RBM) \cite{hinton2006fast}&Hinton&2006 \\
        \hline
        Free Energy Predictive Coding \cite{friston2009predictive}&Friston&2009 \\
        \hline
        BC-DIM Predictive Coding \cite{spratling2016predictive}&Spratling&2009 \\
        \hline
        Predictive Sparse Decomposition \cite{kavukcuoglu2010fast}&Kavukcuoglu&2010 \\
        \hline
        Stacked Denoising Auto-encoders \cite{vincent2010stacked}&Vincent&2010 \\
        \hline
        Deep Predictive Coding Networks \cite{chalasani2013deep}&Chalasani&2013 \\
        \hline
        PredNet \cite{lotter2016deep}&Lotter&2016 \\
        \hline
        Multilevel Predictor Estimator \cite{kim2017predictor}&Kim&2017 \\
        \hline
        Deep Predictive Coding \cite{wen2018deep}&Wen&2018 \\
        \hline
        LPCNet \cite{valin2019lpcnet}&Valin&2019 \\     
        \hline
    \end{tabular}
    \end{center}
    
\vspace{5mm}

Despite the extensive literature, there are key advantages of bio-inspired top-down connections that are not currently being realized in existing predictive coding models. These unrealized advantages include (a) high processing efficiency, (b) input type agnosticism and the ability to process multiple input types simultaneously (multi-modal).

\subsubsection*{Friston's models of free-energy agents}

Friston in an impressive series of papers presents models of agents adjusting  their internal states and sampling  the environment to minimize their free-energy, see for example  \cite{Friston}. These models are   differential equation simulations of  how neural network systems in the brain work. They are mathematical in nature, and are based on functions rather than simulations by computer programs using machine learning models (e.g. artificial  neural networks). In this paper we generate software simulations of learning models, rather than presenting differential equation models of plausible neural networks.

\subsection{ Key advantages of our model}

 In this section we discuss these key advantages of our proposed model in more detail.

\subsubsection*{(a) Computational Efficiency}

In the literature reviewed, computational efficiency is rarely measured and presented. In the one case where computational efficiency was commented on \cite{wen2018deep}, the top-down connections at each layer in the neural network resulted in doubling the computational requirements of the predictive coding network, compared to a standard neural network with only bottom-up connections. 

Biologically speaking, top-down connections allow for certain computations to be handled at lower levels of processing (i.e. the thalamus) freeing up the higher levels of processing (i.e. the cortex) \cite{alitto2003corticothalamic}. This is not currently realized in existing predictive coding models, but is achieved here.

\subsubsection*{(b) Input Agnostic / Multi-modal}

Computational processing of sensory input data is an ever expanding field of research, and there are a plethora of software modules addressing specific types of incoming sensory data \cite{pythonpackages}, with a notable focus on auditory \cite{bourlard2012connectionist} and visual processing \cite{bradski2000opencv}. Corticothalamic connections in the cortex, by contrast, are found in modular structures that seem to repeat themselves for visual, auditory and somatosensory sensory inputs \cite{alitto2003corticothalamic}. If the cortex processes visual, auditory and somatosensory data using similar generalized computational units, then can we produce a single generalized computational model that will equally process visual, auditory and somatosensory input data?

Multi-modal deep neural networks have been shown to improve learning \cite{ngiam2011multimodal}, and even fill in the missing information in both images \cite{jaques2017multimodal} and speech data \cite{xue2015learning}. However, current predictive coding models do not demonstrate the ability to process different inputs types (input agnostic), nor do they demonstrate the ability to process multiple input types simultaneously (multi-modal). In the literature reviewed, each publication only demonstrated the use of one input type, either the processing of images (e.g. CIFAR 10, MNIST), audio (e.g. Birdsongs), or text (e.g. WMT17 QE task).

\vspace{5mm}

    \begin{center}
    \begin{tabular}{ |p{8cm}|p{4cm}| }
        \hline
        \multicolumn{2}{|c|}{Predictive coding models} \\
        \hline
        Model Name&Input Type Used\\
        \hline
        Retinal Predictive Coding \cite{srinivasan1982predictive}&Visual \\
        \hline
        Linear Predictive Coding \cite{o1988linear}&N/A (Theoretical) \\
        \hline
        Cortical Predictive Coding \cite{rao1999predictive}&Visual \\
        \hline
        Restricted Boltzmann machine (RBM) \cite{hinton2006fast}&Visual \\
        \hline
        Free Energy Predictive Coding \cite{friston2009predictive}&Auditory \\
        \hline
        BC-DIM Predictive Coding \cite{spratling2016predictive}&Visual \\
        \hline
        Predictive Sparse Decomposition \cite{kavukcuoglu2010fast}&Visual \\
        \hline
        Stacked Denoising Auto-encoders \cite{vincent2010stacked}&Visual \\
        \hline
        Deep Predictive Coding Networks \cite{chalasani2013deep}&Visual \\
        \hline
        PredNet \cite{lotter2016deep}&Visual \\
        \hline
        Multilevel Predictor Estimator \cite{kim2017predictor}&Text \\
        \hline
        Deep Predictive Coding \cite{wen2018deep}&Visual \\
        \hline
        LPCNet \cite{valin2019lpcnet}&Auditory \\    
        \hline
    \end{tabular}
    \end{center}
    
\vspace{5mm}

% =============================================================== %

\vspace{5mm}

\noindent{}The architecture presented here represents an implementation of the core of Andy Clark's paper ``Whatever next?'' \cite{clark2013whatever}. He states (Section 2.3):
\vspace{5mm}
\begin{quote}
``\textit{In the context of bidirectional hierarchical models of brain function, action-oriented predictive processing yields a new account of the complex interplay between top-down and bottom-up influences on perception and action, and perhaps ultimately of the relations between perception, action, and cognition. As noted by Hohwy (\cite{hohwy2013predictive}, p.320) the generative model providing the “top-down” predictions is here doing much of the more traditionally ``perceptual'' work, with the bottom-up driving signals really providing a kind of ongoing feedback on their activity (by fitting, or failing to fit, the cascade of downward-flowing predictions)... Hierarchical predictive coding delivers, that is to say, a processing regime in which context-sensitivity is fundamental and pervasive.}''\\
\end{quote}
Thus what we present here is a key form of top-down causation in the brain \cite{ellis2016can} \cite{ellis2018top}. What is new in this paper is focusing on the role of the thalamus in this process, in accordance with \cite{alitto2003corticothalamic}, thereby providing a computational model that clarifies how different the role of the thalamus is from just adding in another cortical layer. 

% =============================================================== %

\subsection{Structure of the paper}

In Section 2 we describe the architecture of the CTNN. In Section 3 we describe the simulations run and the observations made, and then discuss the results and draw conclusions in Section 4.

% =============================================================== %

\section{CTNN design}

In this section we present the Corticothalamic Neural Network (CTNN), a generalized computational model that can extend to the visual, auditory and somatosensory systems. We cover the key features (Section \ref{sec:key_features}), the simplified model of the cortex (Section \ref{sec:compu_model}), our representation of the function of the thalamus (Section \ref{sec:thalamus}), and lastly the dataset we used to test the functionality of our model (Section \ref{sec:data_set}).

\subsection{Key features of the CTNN}\label{sec:key_features}

The key feature modelled by the CTNN is the way predictive processing by the cortex results in expectations of sensory input that are conveyed to the thalamus by downward projections from the cortex. Predictions of expected data are compared in the thalamus with incoming data; if the error between the predictions and incoming data exceeds a threshold, an error message is sent to the cortex, enabling it to update its predictions. Since no data is sent when the error is below a threshold, there can be major savings in data processing required in the cortex. If the error signal is compatible with occlusion of data, this processing can fill in the occluded data on the basis of past expectations. 

\subsection{Simplified model of the cortex}\label{sec:compu_model}

We model the cortex's ability to classify and reconstruct inputs with an auto-encoder neural network with 3 encoding and 3 decoding layers. The encoding layers model the cortex's ability to classify objects, while the decoding layers model the cortex's ability to reconstruct an image (i.e. predict) from a classification class. The auto-encoder model represents the first cortical processing level of the neocortex, for each sense (e.g. region V1 in visual processing).

\begin{figure}[ht!]
\centering
\includegraphics[scale=0.5]{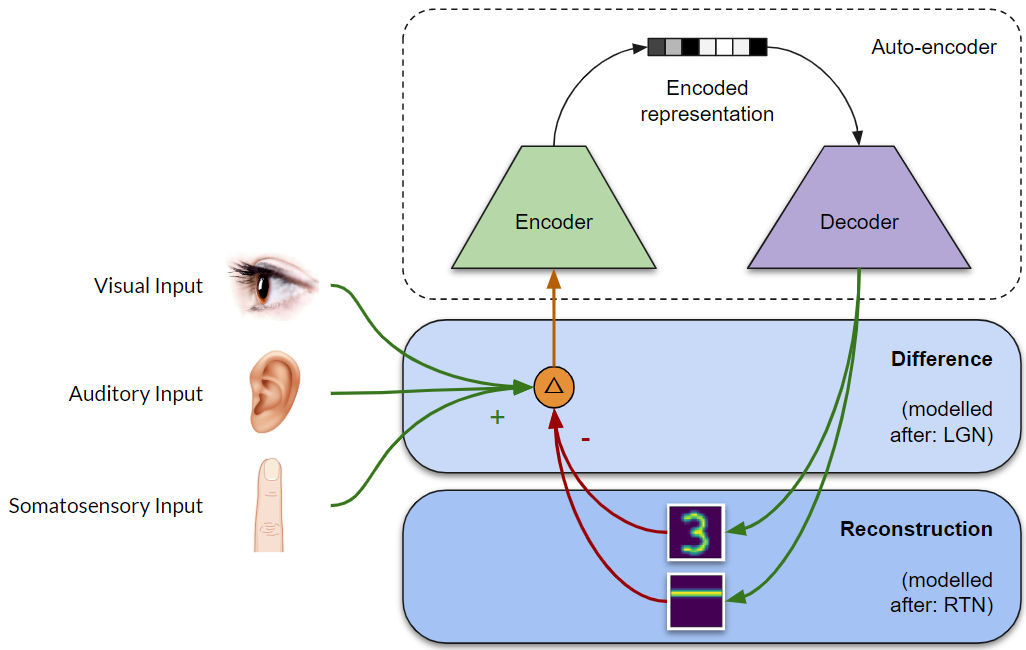}
\caption{\textit{The computational architecture of an auto-encoder model inspired by the corticothalamic connections in the cortex. The reconstruction engine models the reticular nucleus (RTN) in the thalamus, and the difference engine models the part of the thalamus responsible for processing the sensory input (e.g. LGN in the case of the visual sensory system)}}
\label{fig:Fig02}
\end{figure}

The input layer of the auto-encoder consisted of 1,568 nodes (visual 28x28 pixels + audio 28x28 pixels), the encoded layer consisted of 100 nodes, and the output layer of the auto-encoder matched the input layer size: 1,568 nodes (visual 28x28 pixels + audio 28x28 pixels). The output layer of the auto-encoder was connected in series to (1) a reconstruction engine, and then (2) to a difference engine, inspired by the behaviour of the thalamus (see Figure \ref{fig:Fig02}). The reconstruction engine models the reticular nucleus (RTN) in the thalamus, and the difference engine models the part of the thalamus responsible for processing the sensory input (e.g. LGN in the case of the visual sensory system).

\subsection{Simplified model of the thalamus}\label{sec:thalamus}

We model the thalamus as the combination of a reconstruction module and a difference engine. The reconstruction module represents the reticular nucleus (RTN) in the thalamus. In our model, this module converts the 1D output of the decoder (1,568 nodes) into a 2D representation of the visual and audio reconstruction: visual 28x28 pixels + audio 28x28 pixels. This is done so that the output of the auto-encoder could easily be compared to the incoming sensory signal. The output of the reconstruction module is then fed to the difference engine, as an inhibitory signal. 

After the internal representation has been reconstructed, it is compared against the incoming signal by the difference engine. The difference engine takes the pixel-wise difference between the incoming signal (excitatory) and the reconstruction (inhibitory). What set this difference engine apart from other computational models is that it has a built-in threshold mechanism. Inspired by the biological behavior of the thalamus, the difference engine only sends a moderated signal to the auto-encoder if the difference between the incoming sensory signal and the reconstruction exceeds a given internal threshold (see Figure \ref{fig:Fig02}). The difference score ($D$) is calculated as the Mean Square Error (MSE) of the difference between the incoming sensory signal ($y$) and the reconstruction ($\tilde{y}$) (see Equation \ref{eq1} and Equation \ref{eq3}). If the difference score is significant (i.e. it exceeds a given threshold) then the difference engine, modelled after the thalamus, sends the input signal ($y$) to the auto-encoder, modelled after the cortex. Thus the difference score $D$ is defined by
\begin{equation}\label{eq1}
D = \frac{1}{n} \sum_{i=1}^{n} (y_n - \tilde{y}_n \\)^2
\end{equation}
and the output $O$ is defined by
\begin{equation}\label{eq3}
O = 
\begin{cases} 
\text{y} & \text{if } \, D \geq \text{ TH }
\\
0 & \text{if } \, D \, < \text{ TH }
\\
\end{cases}
\end{equation}
where $\text{TH}$ is the threshold.

\vspace*{0.2in}

\subsection{Dataset}\label{sec:data_set}

Our aim is to demonstrate (a) performance on both auditory and visual input types simultaneously, as well as (b) performance over a sequence of similar images. There exist a wide range of audio and visual datasets, commonly used with auto-encoders, including:

\vspace{5mm}

    \begin{center}
    \begin{tabular}{ |p{5cm}|p{7cm}| }
        \hline
        \multicolumn{2}{|c|}{Audio Datasets} \\
        \hline
        Dataset Name&Description\\
        \hline
        MNIST \cite{yann1998mnist}&60,000 images of digits 0-9\\
        \hline
        Spoken Digit Dataset \cite{divyanshu99}&1,500 WAV files of spoken digits 0-9 \\
        \hline
        CUAVE \cite{patterson2002cuave}&36 speakers saying the digits 0-9, in Matlab format\\
        \hline
        AVLetters \cite{matthews2002extraction}&10 speakers saying the letters A to Z, three times each. Raw audio was not available for this dataset\\
        \hline
        AVLetters2 \cite{ngiam2011multimodal}&5 speakers saying the letters A to Z, seven times each. This is a new high definition version of the AVLetters dataset\\
        \hline
        TIMIT \cite{discacoustic}&630 speakers of eight major dialects of American English, each reading ten phonetically rich sentences\\
        \hline
    \end{tabular}
    \end{center}

\vspace{5mm}

\begin{figure}[ht!]
\centering
\includegraphics[scale=0.4]{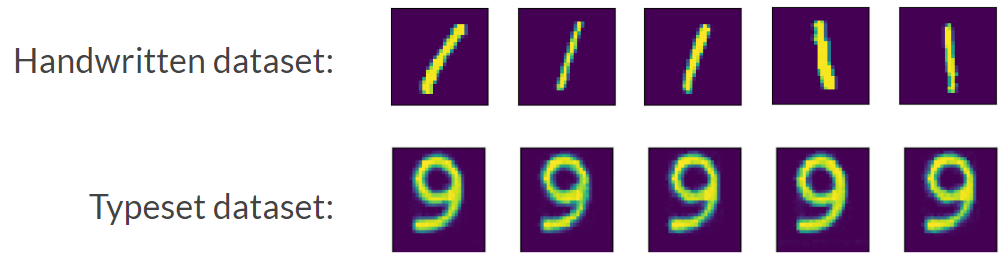}
\caption{\textit{Example of similar handwritten and typeset digits: Handwritten digits of the same class are less similar compared to a typeset dataset.}}
\label{fig:Fig03}
\end{figure}

However, Handwritten digits and spoken words were not similar enough to test for computational efficiency with sequences of similar elements of the dataset, so we chose datasets with high degrees of similarity between elements of the same class (see Figure \ref{fig:Fig03}).

\begin{figure}[ht!]
\centering
\includegraphics[scale=0.5]{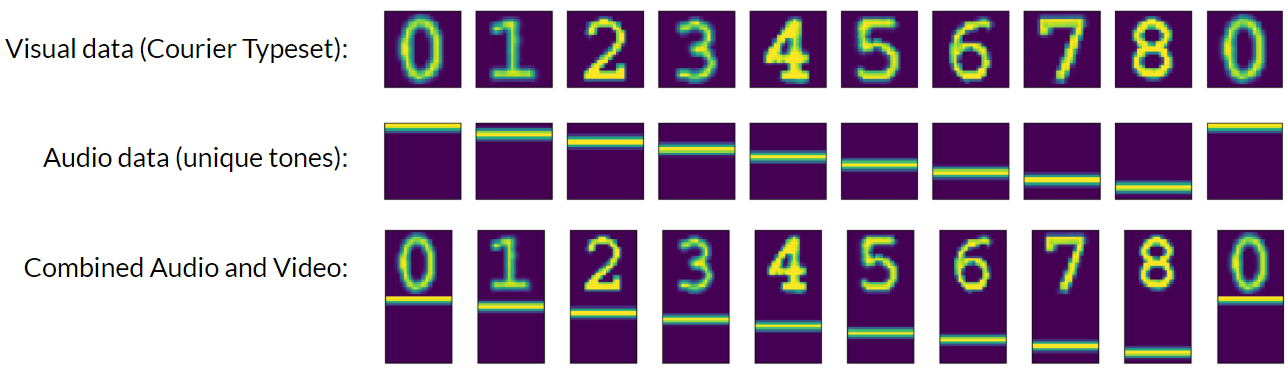}
\caption{\textit{Visual, audio, and combined visual and audio datasets.}}
\label{fig:Fig04}
\end{figure}

For the visual dataset, we use a Courier typeset dataset (digits 0-9) of 28x28 pixels per image \cite{remmelzwaal2019courier}. For the audio dataset, we generated a dataset of tones, where each digit received a unique tone. Each tone was converted into a visual representation of 28x28 pixels, to match the representation of the visual image. The visual and audio representations were then combined into a single input of 1,568 pixels wide (see Figure \ref{fig:Fig04}).

\begin{figure}[ht!]
\centering
\includegraphics[scale=0.6]{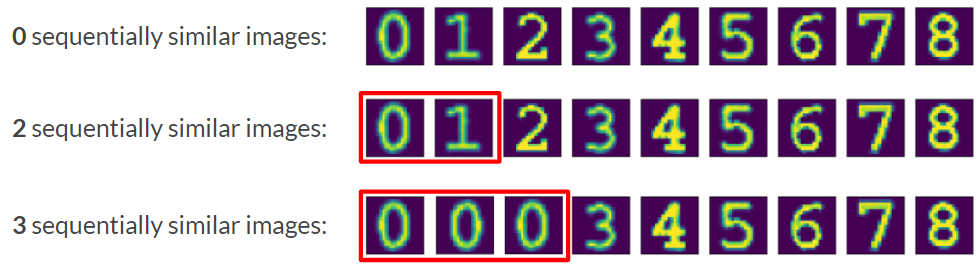}
\caption{\textit{Example of sequences of images containing sequentially similar images}.}
\label{fig:Fig05}
\end{figure}

In addition to the datasets, we dynamically create sequences of similar digits, such that the number of sequentially similar values can be controlled for testing purposes. We avoided sequentially duplicate images, and rather generated images of the same digit class (see Figure \ref{fig:Fig05}).

\vspace{5mm}

% =============================================================== %

\section{Experiments and results}
In this section, we discuss training of the CTNN (Section \ref{sec:training}), and the observations made relating to high processing efficiency (Section \ref{sec:result_high_efficiency}), input agnostic / multi-modal features (Section \ref{sec:result_input_agnostic}) and occlusion robustness (Section \ref{sec:result_occlusion_robust}).

\subsection{Neural network training}\label{sec:training}

\begin{figure}[ht!]
\centering
\includegraphics[scale=0.5]{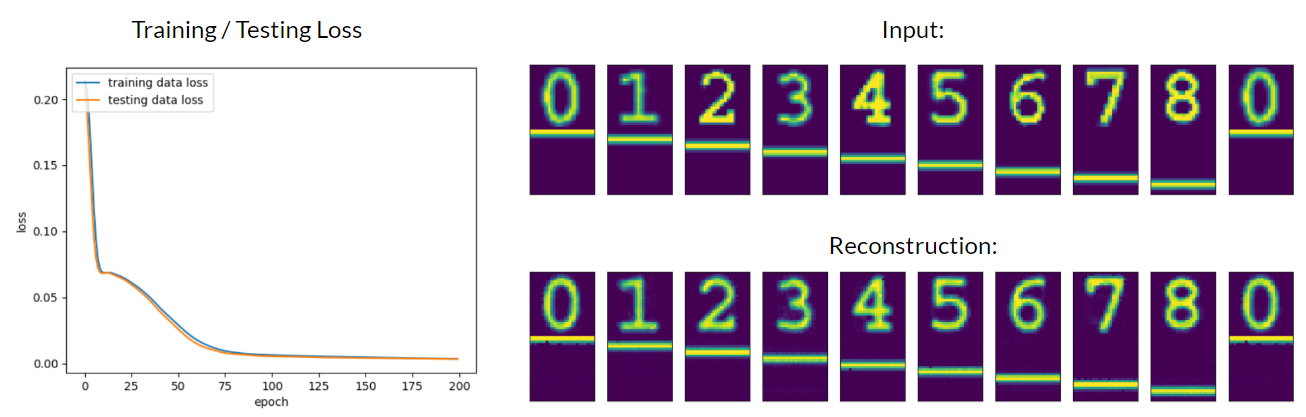}
\caption{\textit{Auto-encoder training and testing losses after 200 epochs}.}
\label{fig:Fig06}
\end{figure}

The auto-encoder was trained on the entire training set of 300 images (30 images x 10 digits) of Courier typeset 0-9. The compile loss function used was the Mean Squared Error (MSE), and training took place over 200 epochs. After 200 epochs, the loss (training data) dropped to 0.0036, and the loss (testing error) dropped to 0.0034, as shown in Figure \ref{fig:Fig06}.

\subsection{Observation 1: High processing efficiency}\label{sec:result_high_efficiency}

\begin{figure}[ht!]
\centering
\includegraphics[scale=0.5]{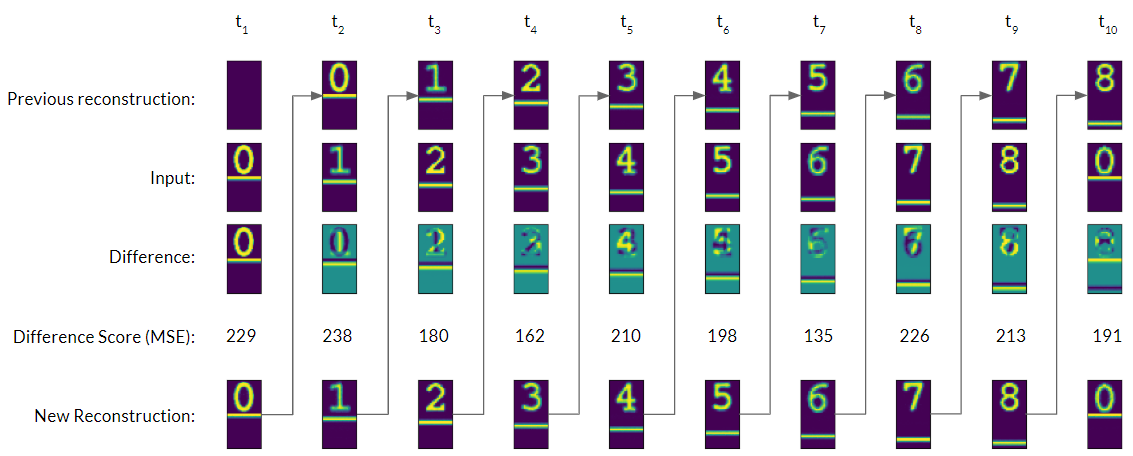}
\caption{\textit{Results from passing the predictive coding model a sequence in inputs containing no sequentially similar images (i.e. no two sequential images belonged to the same class). The first row shows the previous reconstruction, the second row contains the incoming input, the third row the new difference and MSE score, and the last row the new reconstruction.}}
\label{fig:Fig07}
\end{figure}

It was observed that passing a sequence of images to the predictive coding model resulted in a significant difference for each new incoming input sequence. Mathematically, the mean squared error (MSE) exceeded 100 for each image in the sequence (see Figure \ref{fig:Fig07}).

\begin{figure}[ht!]
\centering
\includegraphics[scale=0.5]{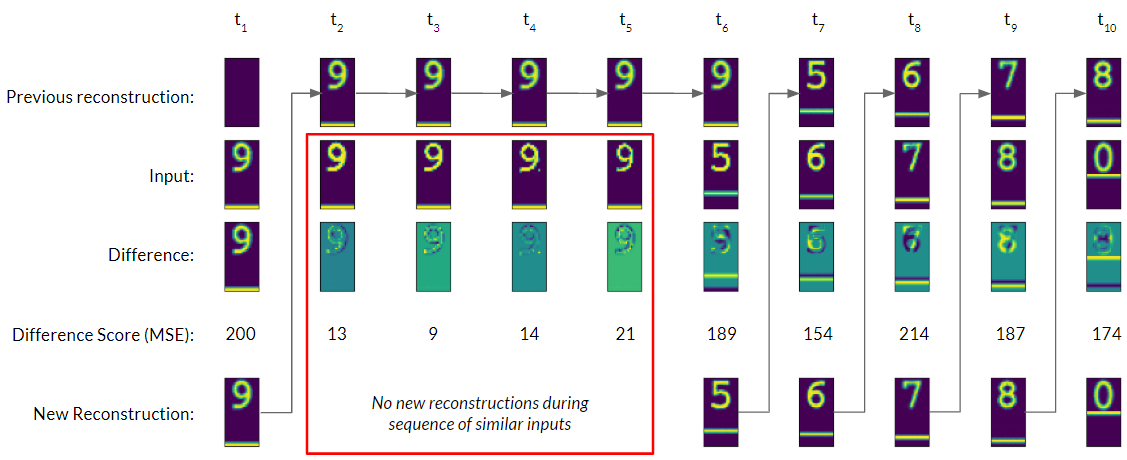}
\caption{\textit{Results from passing the predictive coding model an input sequence with some sequentially similar inputs. The first row shows the previous reconstruction, the second row contains the incoming input, the third row the new difference and MSE score, and the last row the new reconstruction (if a new one is generated)}.}
\label{fig:Fig08}
\end{figure}

We then created an input sequence with some sequentially similar inputs, to observe how similar images affected the processing efficiency. It was observed that passing a sequence of images containing sequential similar images to the predictive coding model resulted in a significantly lower difference for repeated images of the same digit class, compared to images from a difference digit class. Mathematically, the mean squared error (MSE) exceeded 100 for each images of different classes, while the MSE dropped below 20 for similar images in the same digit class (see Figure \ref{fig:Fig08}).

In this model, we set the threshold to 100. Then only if the difference score (MSE) exceeded this threshold value, was the input image sent to the auto-encoder, for the re-construction to be updated. If the difference score (MSE) fell below the threshold value, the auto-encoder was not sent any information, and the reconstruction remained the same. 

From an efficiency perspective, the more times the auto-encoder is engaged the less computationally efficient the model is. For example, if the auto-encoder is required to process every incoming image, then the process requires the full attention of the auto-encoder at a high computational expense. However, if the auto-encoder is not required for a specific sequence, then we achieve higher overall computational efficiency, and the process behaves in a way that resembles `autonomous' behaviour.

\begin{figure}[ht!]
\centering
\includegraphics[scale=0.6]{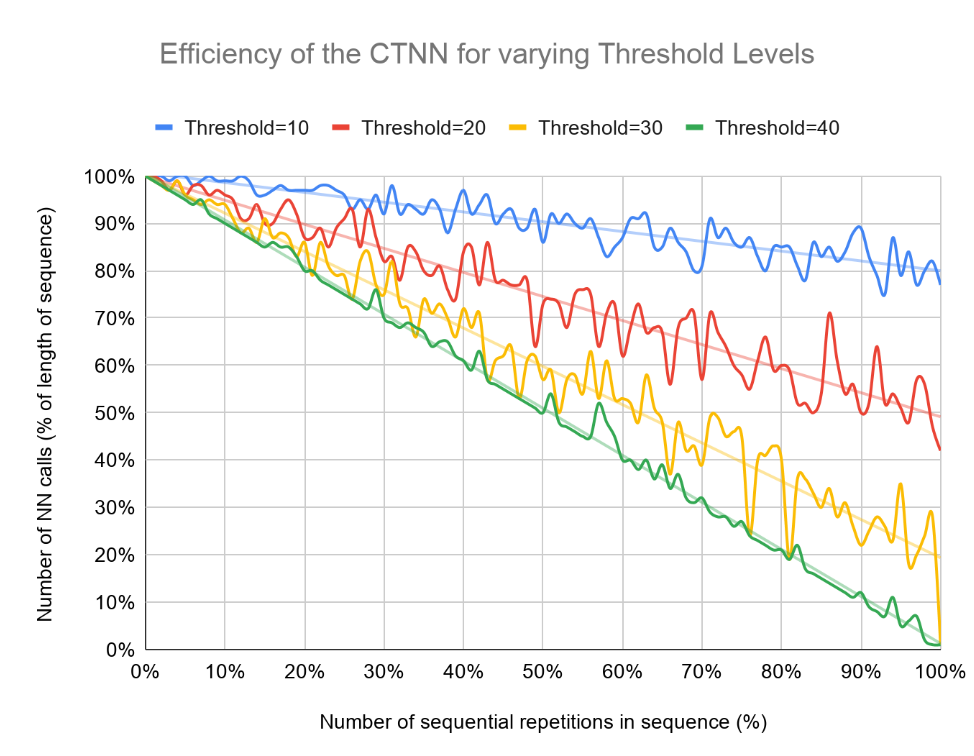}
\caption{\textit{Number of neural network calls in the CTNN model for a varying number of sequentially similar images in a sequence, for different threshold values. The results show that the computational efficiency of the predictive coding model is directly proportional to the number of similar images in the sequence, and that the higher the threshold value, the fewer calls made to the neural network for a similar sequence.}}
\label{fig:Fig09}
\end{figure}

The number of sequentially similar values in a sequence was then varied, and the number of time the auto-encoder was passed an image was plotted against the percentage of sequentially similar values in the sequence (see Figure \ref{fig:Fig09}).

The results from this experiment show that the computational efficiency of the CTNN is directly proportional to the number of similar images in the sequence. From these results we can conclude that thalamus-inspired architecture can reduce overall computational requirement in a predictive coding model proportional to the number of sequentially similar inputs in the dataset. In practice this means that when a sequence of inputs appear similar (e.g. driving down a straight road), the neural network would have reduced involvement, simulating `autonomous' activity.

\subsection{Observation 2: Input Agnostic / Multi-modal}\label{sec:result_input_agnostic}

In this paper we have presented a predictive coding model can accept both visual and auditory inputs (input agnostic), as well as a combined visual and auditory input (multi-modal). This could be extended to accept other input types (e.g. pressure, touch, temperature), depending on the pre-processing of the input signals.

\subsection{Observation 3: Occlusion Robustness}\label{sec:result_occlusion_robust}

\begin{figure}[ht!]
\centering
\includegraphics[scale=0.5]{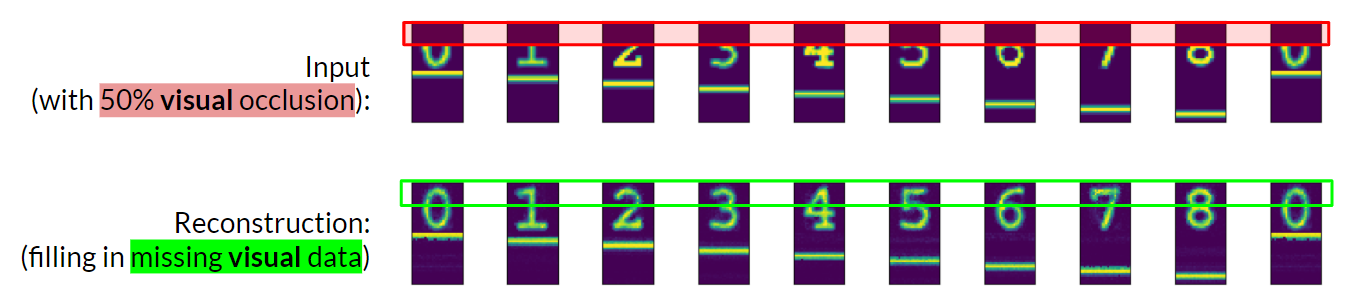}
\caption{\textit{Visual occlusion test: The input and reconstructions when 50\% of the visual input is occluded}.}
\label{fig:Fig10}
\end{figure}

It was anticipated that the CTNN model could demonstrate robustness with respect to input occlusions of one or more sense. To test this we presented the CTNN model with a visual and auditory input, where 50\% of the visual input was occluded (see Figure \ref{fig:Fig10}). The visual occlusion test demonstrated that the predictive coding model was able to successfully reconstruct the missing visual data. 

\begin{figure}[ht!]
\centering
\includegraphics[scale=0.5]{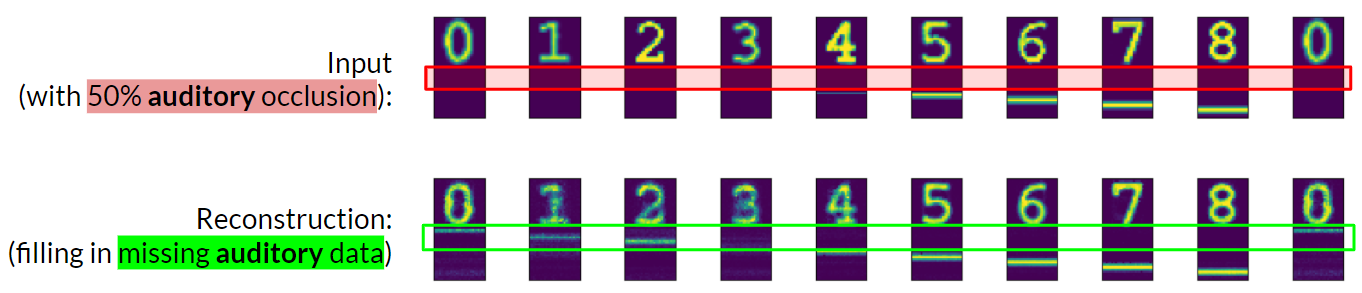}
\caption{\textit{Auditory occlusion test: The input and reconstructions when 50\% of the auditory input is occluded.}}
\label{fig:Fig11}
\end{figure}

The same test was conducted by occluding 50\% of the auditory data, and the test again demonstrated that the predictive coding model was able to successfully reconstruct the missing auditory data (see Figure \ref{fig:Fig11}).

\begin{figure}[ht!]
\centering
\includegraphics[scale=0.7]{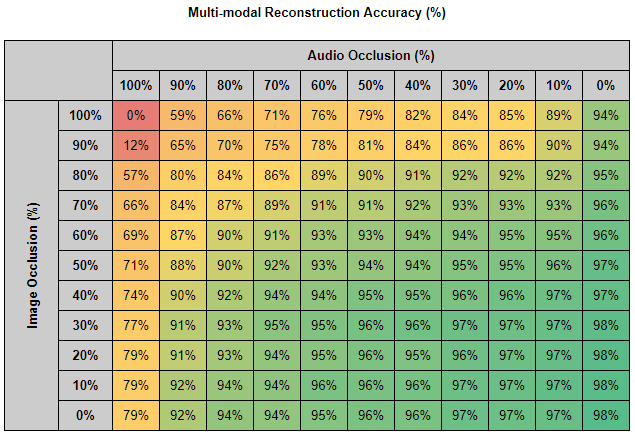}
\caption{\textit{Multi-modal reconstruction accuracy: The reconstruction accuracy (calculated from the normalized MSE score of the difference between the correct reconstruction and the occluded reconstruction), for varying occlusion values for both visual and auditory inputs}}
\label{fig:Fig12}
\end{figure}

To conduct a comprehensive test, the percentage occlusion of both the visual and auditory inputs were varied. It was expected that at low occlusion rates for a single input type (e.g. visual), the reconstruction would demonstrate a high level of robustness. The results were visualized as a heat map (see Figure \ref{fig:Fig12}). The results that the CTNN model is robust to occlusions to individual inputs (e.g. if you can see the number, but cannot hear the tone) up to a significantly high level: only if the visual and auditory sensory inputs were occluded past 70\%, then the reconstruction accuracy dropped below 90\%. This is a remarkable result, demonstrating how a multi-modal model can maintain accuracy well above the occlusion percentage.

% =============================================================== %

\subsection {Altering the threshold}
When the threshold value is altered (see Figure \ref{fig:Fig09}), we find that the higher the threshold value, the fewer the calls made to the neural network for a similar sequence. This is a key point: in broad terms, the level of the threshold corresponds to the level of affect related to the perception \cite{solms2019friston}. If the affective state is high, the threshold is low so the responsitivity is high (keep alert!), and vice versa (a low affective state leads to a high threshold: don't worry!). Thus one only has to do a lot of cortical processing if there is something to worry about. 
This is something we will follow up in a subsequent paper where we combine the CTNN with a SANN \cite{remmelzwaal2020salience}, the latter representing the effect of affect.

% =============================================================== %

\section{Discussion}

In this paper we present a generalized computational model as shown in Figure \ref{fig:Fig02}, inspired by the corticothalamic connections and the behaviour of the thalamus (Figure \ref{fig:Fig13}): the Corticothalamic Neural Network (CTNN). We have demonstrated the following significant features:

\begin{enumerate}
\item The CTNN demonstrates high processing efficiency, specifically when processing sequences of sequentially similar inputs (e.g. images), simulating how autonomous behaviour may arise. Specifically, we have demonstrated that thalamus-inspired architecture reduces overall computational requirement proportional to the number of sequential repetitions in the dataset. This is shown in Section \ref{sec:result_high_efficiency}, and summarised in Figure \ref{fig:Fig09}.
\item The CTNN is input type agnostic and multi-modal: It can accept as inputs various types including auditory, visual and somatosensory, can process multiple input types simultaneously, combining them to an emergent realization, as discussed in Section \ref{sec:result_input_agnostic}. This is a result of the architecture shown in Figure \ref{fig:Fig13}.
\item The CTNN demonstrated a high degree of robustness when there was partial occlusion of one or more sensory input. This is discussed in Section \ref{sec:result_occlusion_robust} and illustrated in Figures \ref{fig:Fig10}, \ref{fig:Fig11}, and \ref{fig:Fig12}.
\end{enumerate}
The CTNN presented in this paper demonstrates how the human cortex may utilise predictive mental representations in processing incoming sensory data \cite{kandel2012age} \cite{frith2013making} \cite{o2014learning}, how corticothalamic connections can reduce the frequency of information sent from lower levels of processing to the higher level of processing in the cortex, and how corticothalamic connections may reduce processing at higher levels of processing in the cortex. \\

The corticothalamic circuitry described by Allito and Usrey \cite{alitto2003corticothalamic} is the result of millions of years of optimisation via Darwinian evolutionary processes; it must have major functional advantages over other possibilities. Guyton and Hall \cite{Guyton}  (page 544) state it this way:
\begin{quote}
``\textit{One of the most important functions of the nervous system
is to process incoming information in such a way that
appropriate mental and motor responses will occur. More
than 99 percent of all sensory information is discarded by
the brain as irrelevant and unimportant. For instance,
one is ordinarily unaware of the parts of the body that
are in contact with clothing, as well as of the seat pres
sure
when sitting. Likewise, attention is drawn only to an
occasional object in one’s field of vision, and even the perpetual
noise of our surroundings is usually relegated to
the subconscious.
But, when important sensory information excites the
mind, it is immediately channeled into proper integrative
and motor regions of the brain to cause desired responses.''}
%This channeling and processing of information is calledthe integrative function of the nervous system.
\end{quote}
The CTNN presented in this paper is a first step towards implementing that function in much more complex ANNs.\\

\subsection{Future work}

Corticothalamic circuitry is very complex \cite{alitto2003corticothalamic} \cite{miller2002spectrotemporal} \cite{sillito2002corticothalamic} \cite{jones2002thalamic} \cite{de2014thalamic} \cite{wimmer2015thalamic} and we have focused only on one element in that function; but it is indeed a key element. Suggestions for future work include:

\begin{enumerate}
 \item Extending the CTNN model to a recurrent neural network, allowing it to process time-series sequences of images.
 \item Extending the input agnostic model presented here, to accept other heterogeneous sensors such as sonar, infra-red, pressure or temperature.
 \item Exploring whether the CTNN architecture might assist in rectifying some of the problems of mis-classification currently occurring in Deep Learning models \cite{Problems}.
 \item Applying the CTNN model to the domain of augmented reality (AR) \cite{AR}, because AR typically requires a subsystem like the thalamus to generate renderings, without which images will always be grainy and sluggish unless supported by massive computational power and data throughput.
 \item Extending our model of the cortex (currently a 6-layer auto-encoder) to use deep neural network architectures \cite{goodfellow2016deep} \cite{Deep_Learning_wason2018deep}.
 \item Extending applications of the CTNN to include contextual linguistic or text recognition, a key application which will require a deep neural network.
 \item Extended the CTNN model to include the features of a \textit{Salience Affected Neural Network}, as presented in \cite{remmelzwaal2020salience}. This will include another key feature of the way the human brain works: namely emotion (affect) plays a key role in intellectual functioning \cite{damasio1994descartes} \cite{panksepp2004affective} \cite{ellis2005neural} \cite{ellis2013affective} \cite{edelman1987neural} \cite{edelman1993neural}. Thus this adds another dimension to the neural processes which nature has found important enough to hard-wire into our brains \cite{ellis2017beyond}. A key feature will be testing different relations between level of affect and threshold size.
\end{enumerate}

\subsection{Supporting Material}

The source code for the CTNN, as well as records of the tests conducted in this paper are publicly available online \cite{remmelzwaal2019ctnn}. For additional information, please contact the corresponding author.

% =============================================================== %

\section*{Acknowledgements}

\noindent{This research did not receive any specific grant from funding agencies in the public, commercial, or not-for-profit sectors.}

% =============================================================== %
%\newpage


\begin{thebibliography}{100}

\bibitem{clark2015embodied} Andy Clark.
\newblock {\em Embodied prediction}.
\newblock Open MIND. Frankfurt am Main: MIND Group, 2015.

\bibitem{alitto2003corticothalamic} Henry~J Alitto and W~Martin Usrey.
\newblock Corticothalamic feedback and sensory processing.
\newblock {\em Current opinion in neurobiology}, 13(4):440--445, 2003.

\bibitem{mountcastle1997columnar} Vernon~B Mountcastle.
\newblock The columnar organization of the neocortex.
\newblock {\em Brain: a journal of neurology}, 120(4):701--722, 1997.

\bibitem{clark2013whatever} Andy Clark.
\newblock Whatever next? predictive brains, situated agents, and the future of
  cognitive science.
\newblock {\em Behavioral and brain sciences}, 36(3):181--204, 2013.

\bibitem{burnod1990adaptive} Yves Burnod.
\newblock {\em An adaptive neural network: the cerebral cortex}.
\newblock Masson editeur, 1990.

\bibitem{miller2002spectrotemporal} Lee~M Miller, Monty~A Escab{\'\i}, Heather~L Read, and Christoph~E Schreiner.
\newblock Spectrotemporal receptive fields in the lemniscal auditory thalamus
  and cortex.
\newblock {\em Journal of neurophysiology}, 87(1):516--527, 2002.

\bibitem{sillito2002corticothalamic} Adam~M Sillito and Helen~E Jones.
\newblock Corticothalamic interactions in the transfer of visual information.
\newblock {\em Philosophical Transactions of the Royal Society of London B:
  Biological Sciences}, 357(1428):1739--1752, 2002.

\bibitem{briggs2008emerging} Farran Briggs and W~Martin Usrey.
\newblock Emerging views of corticothalamic function.
\newblock {\em Current opinion in neurobiology}, 18(4):403--407, 2008.

\bibitem{sherman2009exploring} S~Murray Sherman and RW~Guillery.
\newblock Exploring the thalamus and its role in cortical function. mit press.
\newblock {\em Cambridge, MA}, 2009.

\bibitem{ellis2018top} George Ellis.
\newblock Top-down effects in the brain.
\newblock {\em Physics of life reviews}, 2018.

\bibitem{jones2002thalamic} Edward~G Jones.
\newblock Thalamic circuitry and thalamocortical synchrony.
\newblock {\em Philosophical Transactions of the Royal Society of London B:
  Biological Sciences}, 357(1428):1659--1673, 2002.

\bibitem{shipp2004brain} Stewart Shipp.
\newblock The brain circuitry of attention.
\newblock {\em Trends in cognitive sciences}, 8(5):223--230, 2004.

\bibitem{de2014thalamic} Jos{\'e} de~Bourbon-Teles, Paul Bentley, Saori Koshino, Kushal Shah, Agneish Dutta, Paresh Malhotra, Tobias Egner, Masud Husain, and David Soto.
\newblock Thalamic control of human attention driven by memory and learning.
\newblock {\em Current biology}, 24(9):993--999, 2014.

\bibitem{wimmer2015thalamic} Ralf~D Wimmer, L~Ian Schmitt, Thomas~J Davidson, Miho Nakajima, Karl Deisseroth, and Michael~M Halassa.
\newblock Thalamic control of sensory selection in divided attention.
\newblock {\em Nature}, 526(7575):705, 2015.

\bibitem{friston2018does} Karl Friston.
\newblock Does predictive coding have a future?
\newblock {\em Nature neuroscience}, 21(8):1019, 2018.

\bibitem{thalamic_saalmann2014neura_attentionl} Yuri~B Saalmann and Sabine Kastner.
\newblock Neural mechanisms of spatial attention in the visual thalamus.
\newblock {\em The Oxford Handbook of Attention}, page 399, 2014.

\bibitem{thalamic_control_attention} Jos{\'e} de~Bourbon-Teles, Paul Bentley, Saori Koshino, Kushal Shah, Agneish Dutta, Paresh Malhotra, Tobias Egner, Masud Husain, and David Soto.
\newblock Thalamic control of human attention driven by memory and learning.
\newblock {\em Current biology}, 24(9):993--999, 2014.

\bibitem{thalamic_zikopoulos2007circuits} Basilis Zikopoulos and Helen Barbas.
\newblock Circuits for multisensory integration and attentional modulation through the prefrontal cortex and the thalamic reticular nucleus in primates.
\newblock {\em Reviews in the neurosciences}, 18(6):417--438, 2007.

\bibitem{rees2009visual} Geraint Rees.
\newblock Visual attention: the thalamus at the centre?
\newblock {\em Current biology}, 19(5):R213--R214, 2009.

\bibitem{sun2015human} Lihua Sun, Jari Per{\"a}kyl{\"a}, Markus Polvivaara, Juha {\"O}hman, Jukka Peltola, Kai Lehtim{\"a}ki, Heini Huhtala, and Kaisa~M Hartikainen.
\newblock Human anterior thalamic nuclei are involved in emotion--attention interaction.
\newblock {\em Neuropsychologia}, 78:88--94, 2015.

\bibitem{ellis2017beyond} George Ellis and Mark Solms.
\newblock {\em Beyond Evolutionary Psychology}.
\newblock Cambridge University Press, 2017.

\bibitem{bowman2013attention} Howard Bowman, Marco Filetti, Brad Wyble, and Christian Olivers.
\newblock Attention is more than prediction precision.
\newblock {\em Behavioral and Brain Sciences}, 36(3):206--208, 2013.

\bibitem{purves2010brains} Dale Purves.
\newblock {\em Brains: how they seem to work}.
\newblock Ft Press, 2010.

\bibitem{ungerleider1982two} Leslie~G Ungerleider.
\newblock Two cortical visual systems.
\newblock {\em Analysis of visual behavior}, pages 549--586, 1982.

\bibitem{orban2004comparative} Guy~A Orban, David Van~Essen, and Wim Vanduffel.
\newblock Comparative mapping of higher visual areas in monkeys and humans.
\newblock {\em Trends in cognitive sciences}, 8(7):315--324, 2004.

\bibitem{goodale1992separate} Melvyn~A Goodale and A~David Milner.
\newblock Separate visual pathways for perception and action.
\newblock {\em Trends in neurosciences}, 15(1):20--25, 1992.

\bibitem{hegde2007reappraising} Jay Hegde and Daniel~J Felleman.
\newblock Reappraising the functional implications of the primate visual anatomical hierarchy.
\newblock {\em The Neuroscientist}, 13(5):416--421, 2007.

\bibitem{perry2014feature} Carolyn~Jeane Perry and Mazyar Fallah.
\newblock Feature integration and object representations along the dorsal stream visual hierarchy.
\newblock {\em Frontiers in computational neuroscience}, 8:84, 2014.

\bibitem{hawkins2007intelligence} Jeff Hawkins and Sandra Blakeslee.
\newblock {\em On intelligence: How a new understanding of the brain will lead to the creation of truly intelligent machines}.
\newblock Macmillan, 2007.

\bibitem{sillito2006always} Adam~M Sillito, Javier Cudeiro, and Helen~E Jones.
\newblock Always returning: feedback and sensory processing in visual cortex and thalamus.
\newblock {\em Trends in neurosciences}, 29(6):307--316, 2006.

\bibitem{Cog-Architectures_lieto2018role} Antonio Lieto, Mehul Bhatt, Alessandro Oltramari, and David Vernon.
\newblock The role of cognitive architectures in general artificial intelligence.
\newblock {\em Cognitive Systems Research}, 48:1--3, 2018.

\bibitem{remmelzwaal2020salience} Leendert Remmelzwaal, Jonathan Tapson, and George F R Ellis.
\newblock One-time learning in a biologically-inspired Salience-affected Artificial Neural Network (SANN)
\newblock {\em arXiv preprint arxiv:1908.03532.pdf}, 2020.

\bibitem{edelman2007learning} Gerald~M Edelman.
\newblock Learning in and from brain-based devices.
\newblock {\em Science}, 318(5853):1103--1105, 2007.

\bibitem{nielsen2015neural} Michael~A Nielsen.
\newblock {\em Neural networks and deep learning}, volume~25.
\newblock Determination press USA, 2015.
\newblock \url{http://neuralnetworksanddeeplearning.com/index.html}.

\bibitem{lin2017does} Henry~W Lin, Max Tegmark, and David Rolnick.
\newblock Why does deep and cheap learning work so well?
\newblock {\em Journal of Statistical Physics}, 168(6):1223--1247, 2017.

\bibitem{srinivasan1982predictive} Mandyam~Veerambudi Srinivasan, Simon~Barry Laughlin, and Andreas Dubs.
\newblock Predictive coding: a fresh view of inhibition in the retina.
\newblock {\em Proceedings of the Royal Society of London. Series B. Biological Sciences}, 216(1205):427--459, 1982.

\bibitem{o1988linear} Douglas O'Shaughnessy.
\newblock Linear predictive coding.
\newblock {\em IEEE potentials}, 7(1):29--32, 1988.

\bibitem{rao1999predictive} Rajesh~PN Rao and Dana~H Ballard.
\newblock Predictive coding in the visual cortex: a functional interpretation of some extra-classical receptive-field effects.
\newblock {\em Nature neuroscience}, 2(1):79, 1999.

\bibitem{hinton2006fast} Geoffrey~E Hinton, Simon Osindero, and Yee-Whye Teh.
\newblock A fast learning algorithm for deep belief nets.
\newblock {\em Neural computation}, 18(7):1527--1554, 2006.

\bibitem{friston2009predictive} Karl Friston and Stefan Kiebel.
\newblock Predictive coding under the free-energy principle.
\newblock {\em Philosophical Transactions of the Royal Society B: Biological Sciences}, 364(1521):1211--1221, 2009.

\bibitem{spratling2016predictive} Michael~W Spratling.
\newblock Predictive coding as a model of cognition.
\newblock {\em Cognitive processing}, 17(3):279--305, 2016.

\bibitem{kavukcuoglu2010fast} Koray Kavukcuoglu, Marc'Aurelio Ranzato, and Yann LeCun.
\newblock Fast inference in sparse coding algorithms with applications to object recognition.
\newblock {\em arXiv preprint arXiv:1010.3467}, 2010.

\bibitem{vincent2010stacked} Pascal Vincent, Hugo Larochelle, Isabelle Lajoie, Yoshua Bengio, and
  Pierre-Antoine Manzagol.
\newblock Stacked denoising auto-encoders: Learning useful representations in a
  deep network with a local denoising criterion.
\newblock {\em Journal of machine learning research}, 11(Dec):3371--3408, 2010.

\bibitem{chalasani2013deep} Rakesh Chalasani and Jose~C Principe.
\newblock Deep predictive coding networks.
\newblock {\em arXiv preprint arXiv:1301.3541}, 2013.

\bibitem{lotter2016deep} William Lotter, Gabriel Kreiman, and David Cox.
\newblock Deep predictive coding networks for video prediction and unsupervised learning.
\newblock {\em arXiv preprint arXiv:1605.08104}, 2016.

\bibitem{kim2017predictor} Hyun Kim, Jong-Hyeok Lee, and Seung-Hoon Na.
\newblock Predictor-estimator using multilevel task learning with stack propagation for neural quality estimation.
\newblock In {\em Proceedings of the Second Conference on Machine Translation}, pages 562--568, 2017.

\bibitem{wen2018deep} Haiguang Wen, Kuan Han, Junxing Shi, Yizhen Zhang, Eugenio Culurciello, and Zhongming Liu.
\newblock Deep predictive coding network for object recognition.
\newblock {\em arXiv preprint arXiv:1802.04762}, 2018.

\bibitem{valin2019lpcnet} Jean-Marc Valin and Jan Skoglund.
\newblock Lpcnet: Improving neural speech synthesis through linear prediction.
\newblock In {\em ICASSP 2019-2019 IEEE International Conference on Acoustics, Speech and Signal Processing (ICASSP)}, pages 5891--5895. IEEE, 2019.

\bibitem{Friston}
Karl Friston. 
\newblock A theory of cortical responses. \newblock \textit{Philosophical transactions of the Royal Society B: Biological sciences}, 360:  815-836 (2005).

\bibitem{pythonpackages} Python.org.
\newblock {Python Useful Modules}.
\newblock \url{https://wiki.python.org/moin/UsefulModules}, 2019.
\newblock [Online; accessed 05-January-2019].

\bibitem{bourlard2012connectionist} Herve~A Bourlard and Nelson Morgan.
\newblock {\em Connectionist speech recognition: a hybrid approach}, volume 247.
\newblock Springer Science \& Business Media, 2012.

\bibitem{bradski2000opencv} Gary Bradski and Adrian Kaehler.
\newblock Opencv.
\newblock {\em Dr. Dobb’s journal of software tools}, 3, 2000.

\bibitem{ngiam2011multimodal} Jiquan Ngiam, Aditya Khosla, Mingyu Kim, Juhan Nam, Honglak Lee, and Andrew~Y Ng.
\newblock Multimodal deep learning.
\newblock In {\em Proceedings of the 28th international conference on machine learning (ICML-11)}, pages 689--696, 2011.

\bibitem{jaques2017multimodal} Natasha Jaques, Sara Taylor, Akane Sano, and Rosalind Picard.
\newblock Multimodal auto-encoder: A deep learning approach to filling in missing sensor data and enabling better mood prediction.
\newblock In {\em 2017 Seventh International Conference on Affective Computing and Intelligent Interaction (ACII)}, pages 202--208. IEEE, 2017.

\bibitem{xue2015learning} Wentao Xue, Zhengwei Huang, Xin Luo, and Qirong Mao.
\newblock Learning speech emotion features by joint disentangling-discrimination.
\newblock In {\em 2015 International Conference on Affective Computing and Intelligent Interaction (ACII)}, pages 374--379. IEEE, 2015.

\bibitem{hohwy2013predictive} Jakob Hohwy.
\newblock {\em The predictive mind}.
\newblock Oxford University Press, 2013.

\bibitem{ellis2016can} George Ellis.
\newblock {\em How can Physics Underlie the Mind}.
\newblock Springer, 2016.

\bibitem{yann1998mnist} LeCun Yann, Cortes Corinna, and J~Christopher.
\newblock The mnist database of handwritten digits.
\newblock \url{http://yhann. lecun. com/exdb/mnist}, 1998.

\bibitem{divyanshu99} Spoken Digit Dataset
\newblock {Spoken Digit Dataset}.
\newblock \url{https://www.kaggle.com/divyanshu99/spoken-digit-dataset}, 2018.
\newblock [Online; accessed 13-October-2019].

\bibitem{patterson2002cuave} Eric~K Patterson, Sabri Gurbuz, Zekeriya Tufekci, and John~N Gowdy.
\newblock Cuave: A new audio-visual database for multimodal human-computer interface research.
\newblock In {\em 2002 IEEE International Conference on Acoustics, Speech, and Signal Processing}, volume~2, pages II--2017. IEEE, 2002

\bibitem{matthews2002extraction} Iain Matthews, Timothy~F Cootes, J~Andrew Bangham, Stephen Cox, and Richard Harvey.
\newblock Extraction of visual features for lipreading.
\newblock {\em IEEE Transactions on Pattern Analysis and Machine Intelligence}, 24(2):198--213, 2002.

\bibitem{discacoustic} NIST~Speech Disc, John~S Garofolo, Lori~F Lamel, William~M Fisher, Jonathan~G Fiscus, David~S Pallett, and Nancy~L Dahlgren.
\newblock Acoustic-phonetic continuous speech corpus.

\bibitem{remmelzwaal2019courier} Leendert Remmelzwaal, Gabrielle Coppez, Barry Pitman, Michael Pitman. 
\newblock {Courier Font Dataset}.
\newblock \url{https://bitbucket.org/leenremm/courier_dataset}, 2019.
\newblock [Online; accessed 13-October-2019].

\bibitem{solms2019friston} Mark Solms and  Karl Friston.
\newblock How and why consciousness arises: some considerations from physics and physiology.
\newblock {\em J. Conscious. Stud}. 25:  202-238, 2019.

\bibitem{kandel2012age} Eric~R Kandel.
\newblock {\em The age of insight: The quest to understand the unconscious in
  art, mind, and brain, from Vienna 1900 to the present}.
\newblock Random House Incorporated, 2012.

\bibitem{frith2013making} Chris Frith.
\newblock {\em Making up the mind: How the brain creates our mental world}.
\newblock John Wiley \& Sons, 2013.

\bibitem{o2014learning} Randall~C O'Reilly, Dean Wyatte, and John Rohrlich.
\newblock CTNN: Corticothalamic-inspired neural network
\newblock {\em arXiv preprint arXiv:1407.3432}, 2014.

\bibitem{Guyton}
John E Hall \newblock \textit{Guyton and Hall Textbook of Medical Physiology }
\newblock Saunders Elsevier, 2011.


\bibitem{Problems} Douglas Heaven. 
\newblock Deep Trouble for Deep Learning.
\newblock {\em Nature}, 574:163-166, 2019.

\bibitem{AR} Gary Anthes. 
\newblock Augmented Reality Gets Real. 
\newblock \textit{Communications of the ACM,}  \textbf{62} (9):16-18,  2019.

\bibitem{goodfellow2016deep} Ian Goodfellow, Yoshua Bengio, Aaron Courville, and Yoshua Bengio.
\newblock {\em Deep learning}, volume~1.
\newblock MIT press Cambridge, 2016.

\bibitem{Deep_Learning_wason2018deep} Ritika Wason.
\newblock Deep learning: Evolution and expansion.
\newblock {\em Cognitive Systems Research}, 52:701--708, 2018.

\bibitem{damasio1994descartes} Antonio Damasio.
\newblock {\em Descartes error: Emotion, rationality and the human brain}.
\newblock New York: Putnam, 1994.

\bibitem{panksepp2004affective} Jaak Panksepp.
\newblock {\em Affective neuroscience: The foundations of human and animal emotions}.
\newblock Oxford University Press, 2004.

\bibitem{ellis2005neural} George~FR Ellis and Judith~A Toronchuk.
\newblock Neural development: Affective and immune system influences.
\newblock {\em Consciousness and Emotion: Agency, conscious choice, and selective perception}, 1:81, 2005.

\bibitem{ellis2013affective} Judith~A Toronchuk and George F~R Ellis.
\newblock Affective neuronal selection: the nature of the primordial emotion systems.
\newblock {\em Frontiers in psychology}, 3:589, 2013.

\bibitem{edelman1987neural} Gerald~M Edelman.
\newblock {\em Neural Darwinism: The theory of neuronal group selection.}
\newblock Basic books, 1987.

\bibitem{edelman1993neural} Gerald~M Edelman.
\newblock Neural darwinism: selection and reentrant signaling in higher brain function.
\newblock {\em Neuron}, 10(2):115--125, 1993.

\bibitem{remmelzwaal2019ctnn} Leendert Remmelzwaal, Amit Mishra, George F R Ellis.
\newblock {CTNN Source Code}.
\newblock \url{https://bitbucket.org/leenremm/ctnn}, 2019.

\end{thebibliography}
\end{document}